\documentclass[sigconf, nonacm]{acmart}
\usepackage[ruled,vlined]{algorithm2e}
\RestyleAlgo{ruled}
\RestyleAlgo{vlined}

\def\figurePath{fig/}
\def\myfigure#1#2#3{
\begin{figure}[t]\centering\includegraphics[width = \linewidth]{\figurePath#2}\caption{#3}\label{fig:#1}
\end{figure}}

\def\mycfigure#1#2#3{\begin{figure*}[htb]\centering\includegraphics*[clip, width = \linewidth]{\figurePath#2}\caption{#3}\label{fig:#1}\end{figure*}}

\usepackage{soul}
\usepackage{mathtools} 
\usepackage{xspace} 
\usepackage{multirow}

\copyrightyear{2024}
\acmYear{2024}
\setcopyright{acmlicensed}\acmConference[SIGGRAPH Conference Papers '24]{Special Interest Group on Computer Graphics and Interactive Techniques Conference Conference Papers '24}{July 27-August 1, 2024}{Denver, CO, USA}
\acmBooktitle{Special Interest Group on Computer Graphics and Interactive Techniques Conference Conference Papers '24 (SIGGRAPH Conference Papers '24), July 27-August 1, 2024, Denver, CO, USA}
\acmDOI{10.1145/3641519.3657496}
\acmISBN{979-8-4007-0525-0/24/07}

\begin{document}

\title{Real-time Neural Woven Fabric Rendering}

\author{Xiang Chen}
\orcid{0009-0004-1029-9197}
\affiliation{
 \institution{School of Software, Shandong University}
 \country{China}
}
\email{xiang_chen@mail.sdu.edu.cn}

\author{Lu Wang}
\authornotemark[1]
\orcid{0000-0002-2248-3328}
\affiliation{
 \institution{School of Software, Shandong University}
 \country{China}
}
\email{luwang_hcivr@sdu.edu.cn}

\author{Beibei Wang}
\orcid{0000-0001-8943-8364}
\authornote{Corresponding authors.}
\affiliation{
 \institution{School of Intelligence Science and Technology, Nanjing University}
 \country{China}
}
\email{beibei.wang@nju.edu.cn}

\renewcommand{\shortauthors}{Chen et al.}

\begin{teaserfigure}
  \includegraphics[width=\textwidth]{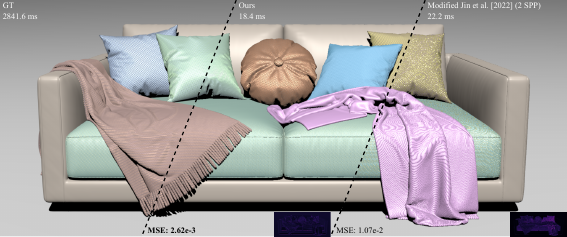}
  \caption{We present a neural network for real-time woven fabric rendering. In this Sofa scene, we provide eight different patterns of woven fabrics. Our method can represent several typical types of woven fabrics with a single neural network, which achieves a fast speed and a quality close to the ground truth}.
  \label{fig:teaser}
\end{teaserfigure}

\begin{abstract}
Woven fabrics are widely used in applications of realistic rendering, where real-time capability is also essential. However, rendering realistic woven fabrics in real time is challenging due to their complex structure and optical appearance, which cause aliasing and noise without many samples. The core of this issue is a multi-scale representation of the fabric shading model, which allows for a fast range query. Some previous neural methods deal with the issue at the cost of training on each material, which limits their practicality. In this paper, we propose a lightweight neural network to represent different types of woven fabrics at different scales. Thanks to the regularity and repetitiveness of woven fabric patterns, our network can encode fabric patterns and parameters as a small latent vector, which is later interpreted by a small decoder, enabling the representation of different types of fabrics. By applying the pixel's footprint as input, our network achieves multi-scale representation. Moreover, our network is fast and occupies little storage because of its lightweight structure. As a result, our method achieves rendering and editing woven fabrics at nearly 60 frames per second on an RTX 3090, showing a quality close to the ground truth and being free from visible aliasing and noise.
\end{abstract}



\maketitle

\section{Introduction}

Realistic woven fabric rendering is vital to many applications in the modern digital age, including virtual reality, video games, and digital humans. Besides the realism, these applications also require a real-time rendering performance. However, simultaneously achieving realism and low time costs is still challenging, as fabrics have complex microstructures and strong anisotropic appearances, leading to aliasing if under-sampled.

Existing approaches can produce high-fidelity rendering results by modeling woven fabrics with volume representations~\cite{Zhao:2011:fabric, heitz2015SGGX} or modeling the scattering function of each fiber~\cite{Khungurn:2015:matching, Aliaga:2017:textileFiber}. Despite their high realism, they can not be applied to real-time rendering. On the other hand, it is more practical to model the woven fabrics with surface shading models~\cite{IrawanAndMarschner2012, Jin:2022:inverse, Zhu:2023:cloth}. Nevertheless, these methods cannot be applied directly to real-time applications since they need hundreds of samples to avoid aliasing caused by detailed yarn structures. The key to this issue is a multi-scale representation of the fabric surface shading model, which can not be treated linearly as the mipmap of diffuse maps~\cite{Williams1983PyramidalP}. The other line of work resorts to neural networks to create a multi-resolution representation for general materials~\cite{Kuznetsov:2021:NeuMip,Gauthier:2022:MipNet}, which are not specialized for fabrics. Despite their high performance, these methods either need per-material training~\cite{Kuznetsov:2021:NeuMip} or have difficulties supporting fabric parameters~\cite{Gauthier:2022:MipNet}.

In this paper, we propose to represent different types of woven fabrics at different scales with a single neural network. At the core of our method is a tiny neural network capable of representing different types of fabrics while simultaneously allowing real-time rendering and editing. The key to achieving these two goals is based on an important observation: woven fabric patterns have regularity, different from the general materials. 
More specifically, starting from the state-of-the-art fabric surface shading model~\cite{Zhu:2023:cloth}, we design a lightweight neural network for the range query of fabric bidirectional scattering distribution functions (BSDFs). Our network first encodes the fabric patterns into features and then fuses the pattern feature with other fabric parameters into a small material latent vector. The material latent vector, together with the queries (position, range, and directions), can be interpreted by a decoder consisting of two multilayer perceptrons (MLPs) into BSDF values. Eventually, the lightweight network structure enables real-time rendering at a frame rate of almost 60 frames per second on an RTX 3090, with a quality close to the ground truth for several typical types of woven fabrics. The network only occupies 5 MB of storage. Furthermore, our network also supports real-time material editing by modifying the fabric patterns and parameters.

\section{Previous work}
\label{sec:previous_work}

In this section, we briefly review previous works related to fabric surface shading models and prefiltering methods for appearance models. As our method is less related to the models based on yarn~\cite{Zhu:2023:yarn}, ply~\cite{Montazeri:2020:ply} or fiber~\cite{Khungurn:2015:matching, Aliaga:2017:textileFiber} curves, we do not discuss these works in this section. The volume-based models~\cite{Zhao:2011:fabric, heitz2015SGGX} are also beyond our scope.

\paragraph{Fabric surface models.}
Fabric surface models~\cite{Adabala:2003:cloth, IrawanAndMarschner2012, Sadeghi:2013:Cloth} have been introduced to fabric rendering for decades. The recent SpongeCake surface shading model by Wang et al.~\shortcite{Wang:2021:Sponge} can model the fabric appearance by stacking different layers, where each layer is defined by a microflake distribution \cite{Jakob:2010:microflake, heitz2015SGGX}. On top of the SpongeCake model, Jin et al.~\shortcite{Jin:2022:inverse} design extra components specialized for fabrics, including a blended Lambertian term and two types of noise. Later, their method is improved by Zhu et al.~\shortcite{Zhu:2023:cloth} to deal with transmission and shadowing-masking effects.

While the above surface shading models can render fabrics efficiently, they are unsuitable for real-time rendering due to the lack of a multi-scale representation. Thus, we propose a neural multi-scale representation that considers range queries and suits real-time rendering. Our method can exploit most of the above models in theory, and we use the modification of Jin et al.~\shortcite{Jin:2022:inverse} (including transmission and shadowing-masking effects) as an example.

\paragraph{Complex appearance prefiltering.}
Due to the limited time budget, prefiltering or downsampling complex appearance models is a practical solution for real-time rendering applications. It has been used for filtering normal maps~\cite{OlanoAndBaker:2010:Lean,Dupuy:2013:Leadar}, filtering the normal distribution function in the slope domain~\cite{Kaplanyan:2016:filtering}, or joint prefiltering~\cite{Xu:2017:LinearMip} the normal map together with bidirectional reflectance distribution function (BRDF). While these approaches produce reasonable results for surface-like distributions, they are unsuitable for fiber-like distributions. Unlike the above methods, Wu et al.~\shortcite{wu2019accurate} introduce a more accurate way to filter displacement maps and BRDFs, using a scaling function for both surface location and direction, at the cost of expensive rendering time. Prefiltering has also been applied to microflake volumes~\cite{heitz2015SGGX, LoubetAndNeyret:2017:LoD,Zhao:2016:Anisotropic}. One related work by Heitz et al.~\shortcite{heitz2015SGGX} performs mipmapping on the matrix which defines the microflake. Despite its efficiency, their method produces less satisfying results, especially for low-roughness fabrics.

Recently, neural-based approaches have been proposed for multi-scale appearance representation. NeuMIP~\cite{Kuznetsov:2021:NeuMip} prefilters the neural textures with a latent texture mipmap. A similar prefiltering idea has been used by Zeltner et al.~\shortcite{Zeltner2023RealTimeNA}, except they encode material parameters into latent vectors. These methods can provide high-quality rendering results and acceptable time costs due to their lightweight network structure. However, they must be trained per material, which costs extra time and restricts their practicality. Unlike over-fitted neural representations, MIPNet~\cite{Gauthier:2022:MipNet} produces mip-mapped textures of spatially varying BRDFs (SVBRDFs). MIPNet requires a tensor-based reformulation of specific shading models, making it less flexible to support other models.

\section{Background and Motivation}

Woven fabrics are made of weft and warp yarns, which are interlaced in a particular pattern. Each yarn is built by twisting plies, where each ply aggregates a set of fibers. In this paper, we focus on fabrics with a single ply for each yarn to demonstrate the effectiveness of our network, following previous work~\cite{Jin:2022:inverse}. These different levels of structures can be modeled in several ways: volume, curve, or surface, where both volume and curve-based models are too heavy for real-time rendering. Hence, we opt for surface models. In this section, we recap the related fabric surface models that are close to our method. Later, we show the motivation and formulation of our method.

\subsection{Preliminaries}

\paragraph{Fabric surface model}
Recently, Jin et al.~\shortcite{Jin:2022:inverse} proposed a surface model for woven fabrics, including the geometric and shading models. In their geometric model, each yarn is expressed as a curved cylinder, defined by the twist and inclination angle of the yarns, and the width and length of the yarns. These parameters with the underlying yarn pattern (Fig.~1 in the supplementary) establish the normal and tangent of yarns, forming microstructures on the cloth surface. Their shading model $f_\mathrm{r}$ consists of a specular term $f_\mathrm{r}^\mathrm{s}$ and a diffuse term $f_\mathrm{r}^\mathrm{d}$, where the former is based on the SpongeCake model~\cite{Wang:2021:Sponge} with a fiber-like microflake phase function, and the latter is a blended Lambertian term considering both macro-surface normal $n_{\mathrm{s}}$ and the ply normal $n_\mathrm{p}$:
\begin{equation}
\begin{aligned}
   f_\mathrm{r}(\omega_\mathrm{i}, \omega_\mathrm{o}) &= f_\mathrm{r}^\mathrm{s}(\omega_\mathrm{i}, \omega_\mathrm{o}) + f_\mathrm{r}^\mathrm{d}(\omega_\mathrm{i}, \omega_\mathrm{o}), \\
   f_\mathrm{r}^\mathrm{s}(\omega_\mathrm{i}, \omega_\mathrm{o}) &= \frac{k_\mathrm{s}D(h)G(\omega_\mathrm{i}, \omega_\mathrm{o})}{4\cos\omega_\mathrm{i}\cdot\cos\omega_\mathrm{o}}, \\
   f_\mathrm{r}^\mathrm{d}(\omega_\mathrm{i}, \omega_\mathrm{o}) &= w\frac{k_\mathrm{d}\langle\omega_\mathrm{i}\cdot n_\mathrm{p}\rangle}{\pi\langle\omega_\mathrm{i}\cdot n_\mathrm{s}\rangle} + (1-w)\frac{k_\mathrm{d}}{\pi}, 
\end{aligned}
 \label{eq:brdf}
\end{equation}

where $\omega_\mathrm{i}$ and $\omega_\mathrm{o}$ are the incoming and outgoing directions, respectively. $k_\mathrm{d}$ and $k_\mathrm{s}$ are the diffuse and specular albedo respectively, and $w$ is the weight for blending the two Lambertian terms. The specular term includes the distribution $D$ on the half vector $h$ between $\omega_\mathrm{i}$ and $\omega_\mathrm{o}$, and the attenuation $G$ along the media with density $\rho$ and thickness $T$. They are defined as:
\begin{equation}
\begin{aligned}
    D(h) &= \frac{1}{\pi\alpha q^2}, \mathrm{~where~} q=h^\top S^{-1} h, \\
    G(\omega_\mathrm{i}, \omega_\mathrm{o}) &= \frac{1-e^{-T\rho(\Lambda(\omega_\mathrm{i})+\Lambda(\omega_\mathrm{o}))}}{\Lambda(\omega_\mathrm{i})+\Lambda(\omega_\mathrm{o})},\\
        \Lambda(\omega) &= \frac{\sigma(\omega)}{\cos\omega}, \mathrm{~where~} \sigma(\omega) = \sqrt{\omega^\top S \omega}, \\
\end{aligned}
\label{eq:DG}
\end{equation}
where $\alpha$ represents roughness and $S$ is a symmetric, positive definite $3 \times 3$ matrix. In detail, $S = \mathrm{diag}(1,1,\alpha^2)$ is for a microflake oriented along the ply direction, and other orientations can be achieved by defining $S^{\prime} = R^{\top}SR$ for any $3 \times 3$ rotation matrix $R$. $\sigma$ is the projected area, and $\Lambda$ is the Smith shadowing-masking function. In practice, $T\rho$ is always assigned a value of 2 for fabrics. The height field scaling factor $\beta$ is introduced to adjust the height field of plies, which affects both the normal $n_\mathrm{p}$ and orientation $t_\mathrm{p}$. All the shading model parameters are summarized in Table~\ref{table:parameters}.

\begin{table} [t]
\caption{ Parameters in the shading model of fabric.}
\begin{center}
\begin{tabular}{ c  c } 
  \hline
  Parameter & Definition \\ 
  \hline
  
  $n_\mathrm{p}$ & ply normal \\ 
  
  $t_\mathrm{p}$ & ply orientation \\ 
  
  $x_\mathrm{p}$ & ply position \\
  
  $k_\mathrm{d}$ ($k_\mathrm{d}^\mathrm{warp}$, $k_\mathrm{d}^\mathrm{weft}$) & diffuse albedo (for warp or weft yarns) \\
  
  $k_\mathrm{s}$ ($k_\mathrm{s}^\mathrm{warp}$, $k_\mathrm{s}^\mathrm{weft}$) & specular albedo (for warp or weft yarns) \\
  
  $\alpha$ ($\alpha^\mathrm{warp}$, $\alpha^\mathrm{weft}$) & roughness (for warp or weft yarns) \\
  
  $\beta$ ($\beta^\mathrm{warp}$, $\beta^\mathrm{weft}$) & height field scaling (for warp or weft yarns) \\
  \hline
\end{tabular}
\label{table:parameters}
\end{center}
\end{table}

Later, Zhu et al.~\shortcite{Zhu:2023:cloth} improve the model above by introducing a bidirectional transmission distribution function (BTDF) and the shadowing-masking effects between the yarns, detailed in the supplementary.

\paragraph{Fabric surface BSDF aggregation}
The woven fabric surface is made of microstructures, which are smaller than a single pixel, leading to a sub-pixel appearance. These micro-structures are captured by sampling the rays within a pixel. Unfortunately, it is impossible to have a high sample rate in real-time rendering. Therefore, the core of real-time fabric rendering is a multi-scale representation which allows efficient aggregation. Zhu et al.~\shortcite{Zhu:2023:cloth} propose an aggregation BSDF defined on a patch $\mathcal{P}$ seen from a pixel, which includes visibility terms to enable shadowing-masking effects between yarns:
\begin{equation} \label{eq:aggregation}
\begin{split}
     f_{\mathcal{P}}(\omega_{\mathrm{i}}, \omega_{\mathrm{o}}) = \frac{1}{A_{\mathcal{P}}(\omega_{\mathrm{o}})} \int_{\mathcal{P}} &f_{p}(\omega_{\mathrm{i}}, \omega_{\mathrm{o}}) \langle \omega_{\mathrm{i}} \cdot n_{\mathrm{p}}(p) \rangle \\ &V(x_{\mathrm{p}}(p), \omega_{\mathrm{i}}) A(p, \omega_{\mathrm{o}}) k_{\mathcal{P}}(p) \mathrm{d}p,
\end{split}
\end{equation}
where $p$ is a point within the patch $\mathcal{P}$. A kernel $k_{\mathcal{P}}$ is defined as $\int_{\mathcal{P}} k_{\mathcal{P}}(p) \mathrm{d}p = 1 $ to normalize the area. $V(x_{\mathrm{p}}, \omega)$ is the binary visibility function of position $x_{\mathrm{p}}$ in direction $\omega$ and $A(p, \omega_{\mathrm{o}})$ is the visible projected area along $\omega_{\mathrm{o}}$ defined as:

\begin{equation}
    A(p, \omega_{\mathrm{o}}) = \frac{\langle \omega_{\mathrm{o}} \cdot n_{\mathrm{p}}(p) \rangle}{\langle n_{\mathrm{s}} \cdot n_{\mathrm{p}}(p) \rangle} V(x_{\mathrm{p}}(p), \omega_{\mathrm{o}}),
\end{equation}
where $n_{\mathrm{s}}$ is surface normal and $\frac{1}{\langle n_{\mathrm{s}} \cdot n_{\mathrm{p}}(p) \rangle}$ is the Jacobian $\lvert \frac{\mathrm{d}x_{\mathrm{p}}(p)}{\mathrm{d}p} \rvert$. $A_{\mathcal{P}}(\omega_{\mathrm{o}})$ is the total visible projected area in patch $\mathcal{P}$ along $\omega_{\mathrm{o}}$: 
\begin{equation}
    A_{\mathcal{P}}(\omega_{\mathrm{o}}) = \int_{\mathcal{P}} \frac{\langle \omega_{\mathrm{o}} \cdot n_{\mathrm{p}}(p) \rangle}{\langle n_{\mathrm{s}} \cdot n_{\mathrm{p}}(p) \rangle} V(x_{\mathrm{p}}(p), \omega_{\mathrm{o}}) k_{\mathcal{P}}(p) \mathrm{d}p = \frac{\langle \omega_{\mathrm{o}} \cdot n_{\mathrm{f}}(\mathcal{P}) \rangle}{\langle n_{\mathrm{s}} \cdot n_{\mathrm{f}}(\mathcal{P}) \rangle}, 
\end{equation}
where $n_{\mathrm{f}}(\mathcal{P})$ is the average visible micro-scale normal over the patch $\mathcal{P}$.

\subsection{Motivation}

Zhu et al.~\shortcite{Zhu:2023:cloth} evaluate the integral of Eqn.~(\ref{eq:aggregation}) in the Monte-Carlo manner, leading to variance when under-sampled and preventing it from being used for real-time rendering. 
Therefore, the core problem is to solve Eqn.~(\ref{eq:aggregation}) efficiently on fabric materials to enable real-time rendering and editing. The function defined by Eqn.~(\ref{eq:aggregation}) is high-dimensional, which includes the geometry and appearance parameters. One straightforward way to represent such a high-dimension function is to use a neural network. This neural network has to meet the following requirements: \emph{fast inference}, \emph{capability of representing multiple woven fabrics}, and \emph{editability}. The fast inference allows real-time rendering; the ability to represent multiple typical woven fabrics with a single neural network which avoids per-material training; and the editability enables material editing. 

The main challenge is designing a network to meet these requirements simultaneously. Fast inference requires a lightweight network; however, a lightweight network usually has limited representation ability. Most lightweight neural networks are used for per-material representation~\cite{Kuznetsov:2021:NeuMip, Zeltner2023RealTimeNA}. To this end, our key insight is the characteristics of woven fabrics, which have regular and repetitive patterns. For editability, we procedurally represent the materials with parameters rather than only using spatially-varying maps~\cite{Zhu:2023:cloth}. In this way, the material has more flexibility for editing.

\section{Method}

In this paper, we first reformulate the fabric aggregation shading model (Sec.~\ref{sec:reformulate}). Then, we design a neural network to represent the shading model (Sec.~\ref{sec:network}). Finally, we apply the neural network for real-time rendering and material editing (Sec.~\ref{sec:application}).

\myfigure{distribution}{Distribution_v7.pdf}{Separation of BSDF distributions. We separate the BSDF value into several components, considering specular/diffuse and yarn type (weft/warp). Thanks to this separation, the distributions become much simpler. The BSDF or components are visualized by mapping the incoming directions to the horizontal axis and the outgoing direction to the vertical axis. Here, we use a plain pattern with white color and compute its BSDF/component values by Monte-Carlo point sampling within a patch with the query size set as $205 \times 205$.}

\subsection{Reformulation of fabric shading model}
\label{sec:reformulate}

The aggregation form of the fabric shading model is shown in Eqn.~(\ref{eq:aggregation}), which is essentially a mapping from the geometry and appearance parameters to a BSDF value, given a query of incoming/outgoing directions and a patch ($\omega_{\mathrm{i}}, \omega_{\mathrm{o}}, \mathcal{P}$):
\begin{equation}
    \{ n_{\mathrm{p}}, t_{\mathrm{p}}, x_{\mathrm{p}}, \alpha, \beta, k_\mathrm{d}, k_\mathrm{s}, \omega_{\mathrm{i}}, \omega_{\mathrm{o}}, \mathcal{P} \} \to f_{\mathcal{P}}(\omega_{\mathrm{i}}, \omega_{\mathrm{o}}),
\end{equation}
where $n_{\mathrm{p}}, t_{\mathrm{p}}, x_{\mathrm{p}}$ are geometry parameters and $\alpha, \beta, k_\mathrm{d}, k_\mathrm{s}$ are appearance parameters defined in Table~\ref{table:parameters}. A straightforward way to represent this mapping is using a neural network directly. However, this mapping mixes several distributions: the specular/diffuse distribution and the warp/weft distribution. As these different distributions have high variance, representing such mapping becomes difficult for a lightweight network. Therefore, we separate the mapping function into different components:
\begin{equation}
    \{ n_{\mathrm{p}}, t_{\mathrm{p}}, \alpha, \beta, \omega_{\mathrm{i}}, \omega_{\mathrm{o}}, \mathcal{P} \} \to \{ C_\mathrm{warp}, C_\mathrm{weft}, S_\mathrm{warp}, S_\mathrm{weft} \},
    \label{eq:formulation}
\end{equation}
where $C_\mathrm{warp}, C_\mathrm{weft}, S_\mathrm{warp}, S_\mathrm{weft}$ represent the diffuse or specular term for warp or weft yarn respectively. We remove $k_\mathrm{d}$ and $k_\mathrm{s}$ from the input, as they depend on the yarn types (weft or warp) only. We also remove $x_{\mathrm{p}}$, as it has been implicitly expressed by normal $n_{\mathrm{p}}$ and orientation $t_{\mathrm{p}}$. The final shading function $f_{\mathcal{P}}$ is defined as: 
\begin{equation}
    f_{\mathcal{P}}(\omega_{\mathrm{i}}, \omega_{\mathrm{o}}) = k_{\mathrm{d}}^{\mathrm{warp}}C_\mathrm{warp} + k_{\mathrm{d}}^{\mathrm{weft}}C_\mathrm{weft} + k_{\mathrm{s}}^{\mathrm{warp}}S_\mathrm{warp} + k_{\mathrm{s}}^{\mathrm{weft}}S_\mathrm{weft}.
\end{equation}

The final formulation is a mapping from geometry parameters ($n_{\mathrm{p}}, t_{\mathrm{p}}$), appearance parameters ($\alpha, \beta$) and a query ($\omega_{\mathrm{i}}, \omega_{\mathrm{o}}, \mathcal{P}$) to four values ($C_\mathrm{warp}, C_\mathrm{weft}, S_\mathrm{warp}, S_\mathrm{weft}$). With this new formulation, the mapping has a much simpler distribution, as shown in Fig.~\ref{fig:distribution}. 

\mycfigure{network_whole}{Network_v10.pdf}{The structure of our neural network. Our network consists of an encoder and a decoder. The encoder takes the input of fabric patterns (normal and orientation textures) and parameters (roughness and height field scaling). These inputs are encoded into a material latent vector by the encoder, which consists of a modified ResNet~\cite{He:2016:Resnet} and an MLP. Then, the material latent vector is fused with the spatial query ($\mathcal{P}$) first, concatenated with the angular inputs ($\omega_{\mathrm{i}}, \omega_{\mathrm{o}}$) and then fed to the angular decoder to get four components. Both the encoder and decoder include a residual block composed of two fully connected layers and a skip connection before the last leaky ReLU function.}

\subsection{Neural representation}
\label{sec:network}

With the reformulated aggregated shading model, we now design a neural network to represent the mapping. As we discussed previously, there are three types of inputs: geometry, appearance, and query parameters. The geometry and appearance parameters establish a material, and the query on the same material should be consistent. To this end, we leverage a typical encoder-decoder structure for the aggregated shading model, where the encoder compresses the geometry and appearance parameters into a material latent vector $z$ and the decoder interprets the latent vector into the different components ($C_\mathrm{warp}, C_\mathrm{weft}, S_\mathrm{warp}, S_\mathrm{weft}$), given a query ($\omega_{\mathrm{i}}, \omega_{\mathrm{o}}, \mathcal{P}$). The network structure is shown in Fig.~\ref{fig:network_whole}. 

\paragraph{Encoder}
The encoder compresses the geometry and appearance parameters into a latent vector $z$. The key to designing this encoder is fusing these two types of parameters. We take two types of inputs: geometry textures for the pattern and the procedural appearance parameters for the others. We first encode textures (normal map and orientation map) into features with ResNet~\cite{He:2016:Resnet}. In practice, we use a modified ResNet-18 without the last residual block, as it does not improve the quality while leading to longer training and inference time. Then, we feed the feature together with the appearance parameters into a residual block, which consists of two fully connected layers of size 128, and output a material latent vector of size 64.

\paragraph{Decoder}
With the material latent vector, we decode it into four components given a query. In the query, we have the spatial inputs (shading position and query range of $\mathcal{P}$) and the angular inputs (light and view directions). The naive way to handle these inputs is concatenating all of them together and interpreting them by a decoder at the same time. However, it does not work well since the spatial parameters dominate the main distribution. Also, the output value must be consistent with varying angular parameters given specific spatial parameters. Therefore, we fuse the material latent vector with the spatial parameters and then concatenate it with the angular parameters. In this way, a small decoder is enough to achieve multi-scale representation. In practice, we apply one-blob encoding~\cite{mueller2021realtime} to the position and size of $\mathcal{P}$, extending them from 3 to 24 channels. This encoding aids the network in distinguishing the spatial inputs better. Then, we use two fully connected layers to fuse the material latent vector with the spatial parameters to get a spatial feature. Finally, the spatial feature, together with the angular inputs, is interpreted by a decoder---\emph{angular decoder} to get the final four BSDF components, where the angular decoder consists of five fully connected layers with one skip connection. Note that the angular inputs consist of three channels ($x$, $y$, $z$) for the light direction and two channels ($x$, $y$) for the view direction. The $z$ component for the light direction is needed, as it might be negative for transmission.

\paragraph{Loss functions}
We compute the ground-truth values of the four terms ($C_\mathrm{warp}, C_\mathrm{weft}, S_\mathrm{warp}, S_\mathrm{weft}$) by sampling the integral in Eqn.~(\ref{eq:aggregation}), and then we introduce a specular loss and a diffuse loss. 

Regarding the specular loss, we find that the distribution of specular has a wide range. Therefore, similar to NeuMIP~\cite{Kuznetsov:2021:NeuMip}, we perform a mapping $g(x)=\ln(kx+1)$ on the ground truth values, and then use the mean squared error (MSE) loss:
\begin{equation}
    Loss_{S}(f^\mathrm{pred}_{S},f^\mathrm{gt}_{S}) = \frac{1}{N}\sum_{i=1}^{N}{(f^\mathrm{pred}_{S} - g(f^\mathrm{gt}_{S}))}^2, 
\end{equation}
where $f^\mathrm{pred}_{S}$ and $f^\mathrm{gt}_{S}$ represent the network output and the target, respectively, for both $S_\mathrm{warp}$ and $S_\mathrm{weft}$. $N$ denotes the number of corresponding terms in a training batch. In practice, we find $k=100$ for BRDF and $k=1000$ for BTDF are good choices.

For the diffuse terms $C_\mathrm{warp}$ and $C_\mathrm{weft}$, we also use the MSE loss:
\begin{equation}
    Loss_{C}(f^\mathrm{pred}_{C},f^\mathrm{gt}_{C}) = \frac{1}{N}\sum_{i=1}^{N}{(f^\mathrm{pred}_{C} - f^\mathrm{gt}_{C})}^2.
\end{equation}

Combining the specular and diffuse loss functions together leads to our final loss:

\begin{equation}
\begin{split}
    Loss(f^\mathrm{pred},f^\mathrm{gt}) &= \sum_{I} \lambda_{S} Loss_{S}^{i}(f^\mathrm{pred}_{S_\mathrm{i}},f^\mathrm{gt}_{S_\mathrm{i}}) \\
    &+ \sum_{I} \lambda_{C} Loss_{C}^\mathrm{i}(f^\mathrm{pred}_{C_\mathrm{i}},f^\mathrm{gt}_{C_\mathrm{i}}),
\end{split}
\end{equation}
where $i$ is one of the yarn types $I$ ($\mathrm{warp}$ or $\mathrm{weft}$). $\lambda_{S}$ and $\lambda_{C}$ are the hyperparameters, set as 0.4 and 0.1 in practice.

\subsection{Rendering and editing}
\label{sec:application}

Now, we use our neural network for real-time rendering and material editing. In practice, we integrate our network into Falcor~\cite{Kallweit22Falcor}, and adopt NVIDIA TensorRT for inference.

\paragraph{Rendering}
Before rendering, for all the fabric materials in the scene, we get their material latent vectors with our encoder. During rendering, we shoot a ray to get a shading point at each pixel. Then, we prepare the query ($\omega_{\mathrm{i}}$, $\omega_{\mathrm{o}}$, $\mathcal{P}$) and compute the lighting. Next, we infer the decoder with the query and the material latent vector to get $C_\mathrm{warp}$, $C_\mathrm{weft}$, $S_\mathrm{warp}$ and $S_\mathrm{weft}$. Finally, we compute the BSDF value with these four components and multiply it by the lighting to get the final rendering result. Note that only the decoder needs to be evaluated during rendering.

\paragraph{Editing}
Our method allows editing of the following parameters: albedos ($k_\mathrm{d}, k_\mathrm{s}$), roughness, height field scaling factor, and fabric patterns. To edit albedos, we only need to change color textures, and no encoder inference is needed, as they are not coupled with the network. To edit the other parameters, we need to infer the encoder to update the material latent vector, which costs about 0.5 ms. After getting the material latent vector, the following steps are the same as rendering. Although the encoder needs to be inferred during editing, it costs negligible time. Thus, our method supports real-time editing of fabric materials.

\section{Implementation details}

\subsection{Data preparation}

To generate the synthetic training dataset, we modify the shading model by Jin et al.~\shortcite{Jin:2022:inverse}, including the transmission and the shadowing-masking effects between yarns within a fixed-size patch. The main difference from Zhu et al.~\shortcite{Zhu:2023:cloth} is that their model includes a delta transmission and computes the shadowing-masking term within the pixel's footprint. Note that our network is compatible with most woven fabric surface shading model, and we choose the current shading model for simplicity.

\begin{table} [t]
\caption{Distributions to sample the parameter space of our model. $\mathcal{U}(x, y)$ represents a continuous uniform distribution in the interval $(x, y)$. $\mathcal{V}(X)$ is a discrete uniform random variable on a finite set $X$.}
\begin{center}
\begin{tabular}{ c  c } 
  \hline
  Parameter & Sampling Function \\ 
  \hline
  yarn pattern & $W = \mathcal{V}(\{0, 1, 2, 3, 4, 5, 6\})$ \\ 
  twist angle & $\psi = \mathcal{V}(\{-30, 0, 30\})$ \\
  inclination angle & $u = \mathcal{U}(15, 45)$ \\
  roughness & $\alpha = \mathcal{U}(0.1, 1)$ \\ 
  height field scaling & $\beta = \mathcal{U}(0, 2)$ \\
  footprint position & $p = (\mathcal{U}(0, 1), \mathcal{U}(0, 1))$ \\
  footprint size & $s = \mathcal{U}(\mathcal{V}(\{(0, 1), (1, 5)\}))$ \\
  incoming direction & $\omega_{\mathrm{i}} = (\mathcal{U}(-1, 1), \mathcal{U}(-1, 1))$ \\
  outgoing direction & $\omega_{\mathrm{o}} = (\mathcal{U}(-1, 1), \mathcal{U}(-1, 1))$ \\
  \hline
\end{tabular}
\label{table:sampling}
\end{center}
\end{table}

We sample parameters to generate the dataset with the sampling functions shown in Table~\ref{table:sampling}. Specifically, we choose seven fabric patterns, including plain, twill $3 \times 3$, twill $5 \times 5$, twill $8 \times 8$, satin $5 \times 5$, satin $8 \times 8$ and satin $5 \times 10$. The gap size is set as 0.2, which means the gap takes up 20\% of the yarn. For each pattern, we generate different sets (9 for the plain pattern and 4 for others) of normal and orientation textures by sampling the twist angle $\psi$ and the inclination angle $u$. Then, we uniformly sample the roughness and height field scaling factor, producing 9 sets for the plain pattern and 16 sets for other patterns. In total, we generate 81 plain materials, 192 twill materials, and 192 satin materials.

For each material, we generate queries by sampling the footprints and the incoming/outgoing directions. For the footprint center, we partition the pattern texture into $8 \times 8$ grids and perform uniform sampling at each grid. Then, for each footprint center, we generate different footprint sizes by uniformly locating 10 samples in the range $\left[0, L_t\right]$ and another 10 samples in the range $\left[L_t, 5L_t\right]$, where $L_t$ is the pattern texture size. Then, we sample the $\omega_i$ and $\omega_o$ by stratified sampling the $x$ and $y$ components of both directions in a $8 \times 8$ grid and computing the $z$ component for both BRDF and BTDF. As a result, we have around 2500 valid pairs of directions for each footprint, leading to about 3.2 million queries per material. After establishing the materials and queries, we compute their BSDF components ($C_\mathrm{warp}$, $C_\mathrm{weft}$, $S_\mathrm{warp}$ and $S_\mathrm{weft}$) using 2048 samples to avoid noise. The time cost of data generation is about 22 minutes per material and 165 hours in total.

\subsection{Training details}

Our network is implemented in the PyTorch framework, using the SGD optimizer. In practice, we use a learning rate of $5 \times 10^{-2}$ for all the weights, which decays by 0.2 at the sixth and ninth epoch, and we also add a regularization to all the weights. During training, materials are randomly chosen for each iteration. We uniformly choose about 160 million queries out of the entire data as the training dataset and organize 512 queries as one batch. We train the network for 10 epochs, which took about 4 days on a single NVIDIA RTX 3090 GPU.

\section{Result}
\label{sec:result}

In this section, we evaluate our method and compare it against previous works. We take 256 samples per pixel (SPP) to get the converged results of the modified Jin et al.~\shortcite{Jin:2022:inverse} as the ground truth (GT). In our experiments, we only consider the direct illumination from a point or directional light source.

All the given timings are total time costs measured on a GeForce RTX 3090 GPU, except for the time costs of the BSDF evaluation in Table~2 (supplementary). We use MSE to measure the difference between each method and the ground truth, and visualize the differences with \protect\reflectbox{F}LIP\xspace~\cite{Andersson:2020:FLIP}. The resolutions of all the results are set as $1920 \times 1080$.

\subsection{Comparison against previous work}

In this section, we make comparisons with modified Jin et al.~\shortcite{Jin:2022:inverse} (taken as baseline) and NeuMIP~\cite{Kuznetsov:2021:NeuMip}. For fairness, we use the same-sized decoder for NeuMIP as our angular decoder, and compare our method with NeuMIP only on seen BRDF materials, since NeuMIP must be trained per material and does not support BTDFs. In Fig.~\ref{fig:compare_neumip}, we compare our method with the baseline (1 SPP), the baseline (equal time) and NeuMIP (1 SPP). By comparison, we find that the baseline (1 SPP) produces results with apparent aliasing, even with a higher sample rate. In contrast, our results are free from aliasing and closer to the ground truth. As for NeuMIP, it can produce results with a higher quality than ours in terms of MSE, at the cost of a higher storage and per-material training. In particular, the storage of our network is less than 5 MB, while the storage of NeuMIP is over 50 MB for each neural material. As NeuMIP is trained per material, its storage increases with the number of materials in the scene, while the storage of our method keeps constant. Furthermore, our network can be applied to three typical types of woven fabrics once trained, but NeuMIP have to be trained per material which takes a long time.

In Fig.~\ref{fig:teaser}, we compare our method with the baseline (2 SPP) on a complex scene with multiple fabrics. The result of the baseline has noticeable aliasing, while our result is smooth and closer to the ground truth. Meanwhile, our result has lower MSE than theirs. We provide more comparisons in Fig.~\ref{fig:compare_BRDF} for BRDF and Fig.~\ref{fig:compare_BTDF} for BTDF on unseen materials. In both scenes, our method performs better than the baseline with 1 SPP in terms of MSE. For equal-time comparison, our results have much less noise and aliasing for both BRDF and BTDF, although our results have higher MSE occasionally. More results of unseen materials are provided in the supplementary.

\subsection{Ablation study}

\paragraph{Encoding of footprint} 
We apply one-blob encoding~\cite{mueller2021realtime} to the query footprint's position and size. We validate its impact in Fig.~\ref{fig:ablation}, by comparing the results with and without the one-blob encoding. By comparison, we find that the result with encoding has lower MSE, less artifacts, and a closer specular shape to the GT. This proves that the encoding improves performance.

\paragraph{Spatial fusion}
In our decoder, we perform the spatial query by fusing the material latent vector with the footprint and then perform the angular query. We validate the impact of this spatial fusion in Fig.~\ref{fig:ablation}, by comparing the results of our network with and without the spatial fusion. For fairness, we increase the number of hidden layers in the angular decoder when removing the spatial fusion. We find that the result without {spatial fusion} shows obvious artifacts and higher MSE.

\subsection{Interpolation and editing}

\paragraph{Interpolation}
The material latent vectors produced by the encoder form a material latent space. We study the behavior of the material latent space by interpolating or extrapolating between two given material latent vectors using parameters as the weights. In practice, we investigate the interpolation and extrapolation of two material latent vectors using the roughness (left) and height field scaling factor (right) as the weights in Fig.~\ref{fig:ITP}. Here, two given material latent vectors are obtained by performing the encoder with the parameter (roughness/height field scaling factor) set as 0.2 and 0.8. Then, the material latent vectors at 0.3 or 0.6 are computed by interpolation, while the material latent vector at 0.9 is computed by extrapolation. We compare the results of these interpolated/extrapolated material latent vectors (bottom) with the ones generated by the encoder directly (top). The results show that the interpolation/extrapolation of material latent vectors can produce a smooth transition and match the results rendered with the directly encoded material latent vectors.

\paragraph{Editing}
In Fig.~\ref{fig:editing}, we show the results of material parameter editing, including roughness, height field scaling factor, pattern and albedos. Our method can produce various rendering results when editing these parameters. In particular, our method is able to represent spatial-varying appearances when applying albedo maps, by computing $k_\mathrm{d}$ and $k_\mathrm{s}$ via mipmapping. In the supplementary video, we also show the efficiency of our material editing.

\begin{table} [t]
\caption{ Time cost of our method for each scene in microsecond. ``Others'' refers to the non-neural steps (ray intersection, data transmission and scene shading).} 
\begin{center}
\begin{tabular}{ c  c  c } 
  \hline
  Scene & Inference & Others \\ 
  \hline
  Single Cloth (Fig.~\ref{fig:compare_neumip}) & 14.3 & 2.9 \\ 
  
  Bowl Cloth (Fig.~\ref{fig:compare_BRDF}) & 13.9 & 4.0 \\
  
  Lamp (Fig.~\ref{fig:compare_BTDF}) & 14.4 & 2.9 \\
  
  Sofa (Fig.~\ref{fig:teaser}) & 13.6 & 4.8 \\
  \hline
\end{tabular}
\label{table:time}
\end{center}
\end{table}

\subsection{Performance analysis}
In Table~\ref{table:time}, we analyze the run-time performance of our method and show the breakdown cost of neural and non-neural steps. The neural step is the inference of our decoder, and the non-neural refers to the steps without the network (ray intersection, data transmission, and shading). We do not include the time cost of our encoder, as it is only inferred when initializing or editing materials, which takes only about 0.5 ms. As shown in the table, our method can achieve real-time rendering. Moreover, as shown in the supplementary video, our method keeps a stable frame rate at different scales when zooming in and out. 

\subsection{Discussion and limitations}
\label{sec:limitation}

Our real-time neural network can represent multi-scale appearances of multiple woven fabrics, which avoids training the network per material. However, our method still has some limitations.

\paragraph{Evaluation time} 
We use TensorRT to deploy our neural network, and the efficiency is optimized automatically. Other techniques could be used to accelerate its speed further. One solution is rewriting it by Slang as Zeltner et al.~\shortcite{Zeltner2023RealTimeNA}. Another option is using a small NeuMIP-style texture pyramid of latent codes to make the decoder smaller.

\paragraph{Network representing ability}
Similar to NeuMIP~\cite{Kuznetsov:2021:NeuMip}, our representing ability is still limited by the network. For example, both NeuMIP and our method can not handle very sharp variations, as shown in the twill results of Fig.~\ref{fig:compare_neumip}.

\paragraph{Importance sampling}
Our paper mainly focuses on BSDF evaluation without considering the importance sampling. To enable importance sampling, one solution is predicting the outgoing direction and its probability density function with a neural network, following Zeltner et al.~\shortcite{Zeltner2023RealTimeNA}. 

\paragraph{Energy conservation}
Similar to prior works~\cite{Zeltner2023RealTimeNA}, our method cannot guarantee energy conservation due to the bias of the neural representation. That is to say, more energy might be added, although we have not observed any noticeable issues.

\paragraph{Capability of our fabric model}
We exclude the delta transmission and procedural perturbation in our model, which restricts the realism of the fabric appearance. It is possible to include the delta transmission by getting the gap ratio for a given query with the neural network and then compute the delta transmission through the gap ratio. For the procedural perturbation, combining our method with some real-time procedural glints approaches may be one solution. 

\section{Conclusion and future work}

In this paper, we present a lightweight neural network to represent multiple woven fabric materials at multiple scales. Our key observation is the regular and repetitive characteristics of woven fabric patterns, which enables the possibility of representing multiple materials with a compact latent space. Specifically, we introduce a simple encoder-decoder structure, where the encoder compresses the fabric pattern and other parameters into a material latent vector, and the decoder interprets the material latent vector with a spatial fusion component and a small angular decoder. Thanks to the lightweight encoder and decoder, our network is able to achieve real-time rendering and editing. Meanwhile, our network only occupies a small storage of 5 MB, even for scenes with multiple fabric materials.

In the future, there are still many potential research directions. Currently, our network supports three typical woven fabric types. More types of woven fabrics and knitted fabrics are still missing from our representation. The other types of woven fabrics can be handled by expanding our dataset. However, representing knitted fabrics is difficult, as their structures significantly differ from woven fabrics. We leave it for future work. It is also worth exploring ways to enable rendering fabrics with spatially-varying patterns.

\begin{acks}
We thank the reviewers for the valuable comments. This work has been partially supported by the National Science and Technology Major Project under grant No. 2022ZD0116305 and National Natural Science Foundation of China under grant No. 62272275 and 62172220. 
\end{acks}

\bibliographystyle{ACM-Reference-Format}
\bibliography{paper}


\begin{thebibliography}{28}


\ifx \showCODEN    \undefined \def \showCODEN     #1{\unskip}     \fi
\ifx \showDOI      \undefined \def \showDOI       #1{#1}\fi
\ifx \showISBNx    \undefined \def \showISBNx     #1{\unskip}     \fi
\ifx \showISBNxiii \undefined \def \showISBNxiii  #1{\unskip}     \fi
\ifx \showISSN     \undefined \def \showISSN      #1{\unskip}     \fi
\ifx \showLCCN     \undefined \def \showLCCN      #1{\unskip}     \fi
\ifx \shownote     \undefined \def \shownote      #1{#1}          \fi
\ifx \showarticletitle \undefined \def \showarticletitle #1{#1}   \fi
\ifx \showURL      \undefined \def \showURL       {\relax}        \fi
\providecommand\bibfield[2]{#2}
\providecommand\bibinfo[2]{#2}
\providecommand\natexlab[1]{#1}
\providecommand\showeprint[2][]{arXiv:#2}

\bibitem[\protect\citeauthoryear{Adabala, Magnenat-Thalmann, and Fei}{Adabala et~al\mbox{.}}{2003}]%
        {Adabala:2003:cloth}
\bibfield{author}{\bibinfo{person}{Neeharika Adabala}, \bibinfo{person}{Nadia Magnenat-Thalmann}, {and} \bibinfo{person}{Guangzheng Fei}.} \bibinfo{year}{2003}\natexlab{}.
\newblock \showarticletitle{Real-time rendering of woven clothes}. In \bibinfo{booktitle}{\emph{Proceedings of the ACM Symposium on Virtual Reality Software and Technology}} \emph{(\bibinfo{series}{VRST '03})}. \bibinfo{publisher}{Association for Computing Machinery}, \bibinfo{address}{New York, NY, USA}, \bibinfo{pages}{41–47}.
\newblock
\showISBNx{1581135696}
\urldef\tempurl%
\url{https://doi.org/10.1145/1008653.1008663}
\showDOI{\tempurl}


\bibitem[\protect\citeauthoryear{Aliaga, Castillo, Gutierrez, Otaduy, Lopez-Moreno, and Jarabo}{Aliaga et~al\mbox{.}}{2017}]%
        {Aliaga:2017:textileFiber}
\bibfield{author}{\bibinfo{person}{Carlos Aliaga}, \bibinfo{person}{Carlos Castillo}, \bibinfo{person}{Diego Gutierrez}, \bibinfo{person}{Miguel~A. Otaduy}, \bibinfo{person}{Jorge Lopez-Moreno}, {and} \bibinfo{person}{Adrian Jarabo}.} \bibinfo{year}{2017}\natexlab{}.
\newblock \showarticletitle{An Appearance Model for Textile Fibers}.
\newblock \bibinfo{journal}{\emph{Computer Graphics Forum}} \bibinfo{volume}{36}, \bibinfo{number}{4} (\bibinfo{year}{2017}), \bibinfo{pages}{35--45}.
\newblock
\urldef\tempurl%
\url{https://doi.org/10.1111/cgf.13222}
\showDOI{\tempurl}


\bibitem[\protect\citeauthoryear{Andersson, Nilsson, Shirley, and Akenine{-}M{\"{o}}ller}{Andersson et~al\mbox{.}}{2021}]%
        {Andersson:2020:FLIP}
\bibfield{author}{\bibinfo{person}{Pontus Andersson}, \bibinfo{person}{Jim Nilsson}, \bibinfo{person}{Peter Shirley}, {and} \bibinfo{person}{Tomas Akenine{-}M{\"{o}}ller}.} \bibinfo{year}{2021}\natexlab{}.
\newblock \showarticletitle{{Visualizing Errors in Rendered High Dynamic Range Images}}. In \bibinfo{booktitle}{\emph{Eurographics Short Papers}}.
\newblock
\urldef\tempurl%
\url{https://doi.org/10.2312/egs.20211015}
\showDOI{\tempurl}


\bibitem[\protect\citeauthoryear{Dupuy, Heitz, Iehl, Poulin, Neyret, and Ostromoukhov}{Dupuy et~al\mbox{.}}{2013}]%
        {Dupuy:2013:Leadar}
\bibfield{author}{\bibinfo{person}{Jonathan Dupuy}, \bibinfo{person}{Eric Heitz}, \bibinfo{person}{Jean-Claude Iehl}, \bibinfo{person}{Pierre Poulin}, \bibinfo{person}{Fabrice Neyret}, {and} \bibinfo{person}{Victor Ostromoukhov}.} \bibinfo{year}{2013}\natexlab{}.
\newblock \showarticletitle{Linear Efficient Antialiased Displacement and Reflectance Mapping}.
\newblock \bibinfo{journal}{\emph{ACM Trans. Graph.}} \bibinfo{volume}{32}, \bibinfo{number}{6}, Article \bibinfo{articleno}{211} (\bibinfo{date}{nov} \bibinfo{year}{2013}), \bibinfo{numpages}{11}~pages.
\newblock
\showISSN{0730-0301}
\urldef\tempurl%
\url{https://doi.org/10.1145/2508363.2508422}
\showDOI{\tempurl}


\bibitem[\protect\citeauthoryear{Gauthier, Faury, Levallois, Thonat, Thiery, and Boubekeur}{Gauthier et~al\mbox{.}}{2022}]%
        {Gauthier:2022:MipNet}
\bibfield{author}{\bibinfo{person}{Alban Gauthier}, \bibinfo{person}{Robin Faury}, \bibinfo{person}{J\'{e}r\'{e}my Levallois}, \bibinfo{person}{Th\'{e}o Thonat}, \bibinfo{person}{Jean-Marc Thiery}, {and} \bibinfo{person}{Tamy Boubekeur}.} \bibinfo{year}{2022}\natexlab{}.
\newblock \showarticletitle{MIPNet: Neural Normal-to-Anisotropic-Roughness MIP Mapping}.
\newblock \bibinfo{journal}{\emph{ACM Trans. Graph.}} \bibinfo{volume}{41}, \bibinfo{number}{6}, Article \bibinfo{articleno}{246} (\bibinfo{date}{nov} \bibinfo{year}{2022}), \bibinfo{numpages}{12}~pages.
\newblock
\showISSN{0730-0301}
\urldef\tempurl%
\url{https://doi.org/10.1145/3550454.3555487}
\showDOI{\tempurl}


\bibitem[\protect\citeauthoryear{He, Zhang, Ren, and Sun}{He et~al\mbox{.}}{2016}]%
        {He:2016:Resnet}
\bibfield{author}{\bibinfo{person}{Kaiming He}, \bibinfo{person}{Xiangyu Zhang}, \bibinfo{person}{Shaoqing Ren}, {and} \bibinfo{person}{Jian Sun}.} \bibinfo{year}{2016}\natexlab{}.
\newblock \showarticletitle{Deep Residual Learning for Image Recognition}. In \bibinfo{booktitle}{\emph{2016 IEEE Conference on Computer Vision and Pattern Recognition (CVPR)}}. \bibinfo{pages}{770--778}.
\newblock
\urldef\tempurl%
\url{https://doi.org/10.1109/CVPR.2016.90}
\showDOI{\tempurl}


\bibitem[\protect\citeauthoryear{Heitz, Dupuy, Crassin, and Dachsbacher}{Heitz et~al\mbox{.}}{2015}]%
        {heitz2015SGGX}
\bibfield{author}{\bibinfo{person}{Eric Heitz}, \bibinfo{person}{Jonathan Dupuy}, \bibinfo{person}{Cyril Crassin}, {and} \bibinfo{person}{Carsten Dachsbacher}.} \bibinfo{year}{2015}\natexlab{}.
\newblock \showarticletitle{The SGGX microflake distribution}.
\newblock \bibinfo{journal}{\emph{ACM Trans. Graph.}} \bibinfo{volume}{34}, \bibinfo{number}{4}, Article \bibinfo{articleno}{48} (\bibinfo{date}{jul} \bibinfo{year}{2015}), \bibinfo{numpages}{11}~pages.
\newblock
\showISSN{0730-0301}
\urldef\tempurl%
\url{https://doi.org/10.1145/2766988}
\showDOI{\tempurl}


\bibitem[\protect\citeauthoryear{Irawan and Marschner}{Irawan and Marschner}{2012}]%
        {IrawanAndMarschner2012}
\bibfield{author}{\bibinfo{person}{Piti Irawan} {and} \bibinfo{person}{Steve Marschner}.} \bibinfo{year}{2012}\natexlab{}.
\newblock \showarticletitle{Specular reflection from woven cloth}.
\newblock \bibinfo{journal}{\emph{ACM Trans. Graph.}} \bibinfo{volume}{31}, \bibinfo{number}{1}, Article \bibinfo{articleno}{11} (\bibinfo{date}{feb} \bibinfo{year}{2012}), \bibinfo{numpages}{20}~pages.
\newblock
\showISSN{0730-0301}
\urldef\tempurl%
\url{https://doi.org/10.1145/2077341.2077352}
\showDOI{\tempurl}


\bibitem[\protect\citeauthoryear{Jakob, Arbree, Moon, Bala, and Marschner}{Jakob et~al\mbox{.}}{2010}]%
        {Jakob:2010:microflake}
\bibfield{author}{\bibinfo{person}{Wenzel Jakob}, \bibinfo{person}{Adam Arbree}, \bibinfo{person}{Jonathan~T. Moon}, \bibinfo{person}{Kavita Bala}, {and} \bibinfo{person}{Steve Marschner}.} \bibinfo{year}{2010}\natexlab{}.
\newblock \showarticletitle{A radiative transfer framework for rendering materials with anisotropic structure}.
\newblock \bibinfo{journal}{\emph{ACM Trans. Graph.}} \bibinfo{volume}{29}, \bibinfo{number}{4}, Article \bibinfo{articleno}{53} (\bibinfo{date}{jul} \bibinfo{year}{2010}), \bibinfo{numpages}{13}~pages.
\newblock
\showISSN{0730-0301}
\urldef\tempurl%
\url{https://doi.org/10.1145/1778765.1778790}
\showDOI{\tempurl}


\bibitem[\protect\citeauthoryear{Jin, Wang, Hasan, Guo, Marschner, and Yan}{Jin et~al\mbox{.}}{2022}]%
        {Jin:2022:inverse}
\bibfield{author}{\bibinfo{person}{Wenhua Jin}, \bibinfo{person}{Beibei Wang}, \bibinfo{person}{Milos Hasan}, \bibinfo{person}{Yu Guo}, \bibinfo{person}{Steve Marschner}, {and} \bibinfo{person}{Ling-Qi Yan}.} \bibinfo{year}{2022}\natexlab{}.
\newblock \showarticletitle{Woven Fabric Capture from a Single Photo}. In \bibinfo{booktitle}{\emph{SIGGRAPH Asia 2022 Conference Papers}} \emph{(\bibinfo{series}{SA '22})}. \bibinfo{publisher}{Association for Computing Machinery}, \bibinfo{address}{New York, NY, USA}, Article \bibinfo{articleno}{33}, \bibinfo{numpages}{8}~pages.
\newblock
\showISBNx{9781450394703}
\urldef\tempurl%
\url{https://doi.org/10.1145/3550469.3555380}
\showDOI{\tempurl}


\bibitem[\protect\citeauthoryear{Kallweit, Clarberg, Kolb, Davidovi{\v c}, Yao, Foley, He, Wu, Chen, Akenine-M{\"o}ller, Wyman, Crassin, and Benty}{Kallweit et~al\mbox{.}}{2022}]%
        {Kallweit22Falcor}
\bibfield{author}{\bibinfo{person}{Simon Kallweit}, \bibinfo{person}{Petrik Clarberg}, \bibinfo{person}{Craig Kolb}, \bibinfo{person}{Tom{'a}{\v s} Davidovi{\v c}}, \bibinfo{person}{Kai-Hwa Yao}, \bibinfo{person}{Theresa Foley}, \bibinfo{person}{Yong He}, \bibinfo{person}{Lifan Wu}, \bibinfo{person}{Lucy Chen}, \bibinfo{person}{Tomas Akenine-M{\"o}ller}, \bibinfo{person}{Chris Wyman}, \bibinfo{person}{Cyril Crassin}, {and} \bibinfo{person}{Nir Benty}.} \bibinfo{year}{2022}\natexlab{}.
\newblock \bibinfo{title}{The {Falcor} Rendering Framework}.
\newblock
\newblock
\urldef\tempurl%
\url{https://github.com/NVIDIAGameWorks/Falcor}
\showURL{%
\tempurl}


\bibitem[\protect\citeauthoryear{Kaplanyan, Hill, Patney, and Lefohn}{Kaplanyan et~al\mbox{.}}{2016}]%
        {Kaplanyan:2016:filtering}
\bibfield{author}{\bibinfo{person}{Anton~S. Kaplanyan}, \bibinfo{person}{Stephen Hill}, \bibinfo{person}{Anjul Patney}, {and} \bibinfo{person}{Aaron Lefohn}.} \bibinfo{year}{2016}\natexlab{}.
\newblock \showarticletitle{{Filtering Distributions of Normals for Shading Antialiasing}}. In \bibinfo{booktitle}{\emph{Eurographics/ ACM SIGGRAPH Symposium on High Performance Graphics}}, \bibfield{editor}{\bibinfo{person}{Ulf Assarsson} {and} \bibinfo{person}{Warren Hunt}} (Eds.). \bibinfo{publisher}{The Eurographics Association}.
\newblock
\urldef\tempurl%
\url{https://doi.org/10.2312/hpg.20161201}
\showDOI{\tempurl}


\bibitem[\protect\citeauthoryear{Khungurn, Schroeder, Zhao, Bala, and Marschner}{Khungurn et~al\mbox{.}}{2016}]%
        {Khungurn:2015:matching}
\bibfield{author}{\bibinfo{person}{Pramook Khungurn}, \bibinfo{person}{Daniel Schroeder}, \bibinfo{person}{Shuang Zhao}, \bibinfo{person}{Kavita Bala}, {and} \bibinfo{person}{Steve Marschner}.} \bibinfo{year}{2016}\natexlab{}.
\newblock \showarticletitle{Matching Real Fabrics with Micro-Appearance Models}.
\newblock \bibinfo{journal}{\emph{ACM Trans. Graph.}} \bibinfo{volume}{35}, \bibinfo{number}{1}, Article \bibinfo{articleno}{1} (\bibinfo{date}{dec} \bibinfo{year}{2016}), \bibinfo{numpages}{26}~pages.
\newblock
\showISSN{0730-0301}
\urldef\tempurl%
\url{https://doi.org/10.1145/2818648}
\showDOI{\tempurl}


\bibitem[\protect\citeauthoryear{Kuznetsov, Mullia, Xu, Ha\v{s}an, and Ramamoorthi}{Kuznetsov et~al\mbox{.}}{2021}]%
        {Kuznetsov:2021:NeuMip}
\bibfield{author}{\bibinfo{person}{Alexandr Kuznetsov}, \bibinfo{person}{Krishna Mullia}, \bibinfo{person}{Zexiang Xu}, \bibinfo{person}{Milo\v{s} Ha\v{s}an}, {and} \bibinfo{person}{Ravi Ramamoorthi}.} \bibinfo{year}{2021}\natexlab{}.
\newblock \showarticletitle{NeuMIP: Multi-Resolution Neural Materials}.
\newblock \bibinfo{journal}{\emph{ACM Trans. Graph.}} \bibinfo{volume}{40}, \bibinfo{number}{4}, Article \bibinfo{articleno}{175} (\bibinfo{date}{jul} \bibinfo{year}{2021}), \bibinfo{numpages}{13}~pages.
\newblock
\showISSN{0730-0301}
\urldef\tempurl%
\url{https://doi.org/10.1145/3450626.3459795}
\showDOI{\tempurl}


\bibitem[\protect\citeauthoryear{Loubet and Neyret}{Loubet and Neyret}{2017}]%
        {LoubetAndNeyret:2017:LoD}
\bibfield{author}{\bibinfo{person}{Guillaume Loubet} {and} \bibinfo{person}{Fabrice Neyret}.} \bibinfo{year}{2017}\natexlab{}.
\newblock \showarticletitle{Hybrid mesh-volume LoDs for all-scale pre-filtering of complex 3D assets}.
\newblock \bibinfo{journal}{\emph{Computer Graphics Forum}} \bibinfo{volume}{36}, \bibinfo{number}{2} (\bibinfo{year}{2017}), \bibinfo{pages}{431--442}.
\newblock
\urldef\tempurl%
\url{https://doi.org/10.1111/cgf.13138}
\showDOI{\tempurl}


\bibitem[\protect\citeauthoryear{Montazeri, Gammelmark, Zhao, and Jensen}{Montazeri et~al\mbox{.}}{2020}]%
        {Montazeri:2020:ply}
\bibfield{author}{\bibinfo{person}{Zahra Montazeri}, \bibinfo{person}{S\o{}ren~B. Gammelmark}, \bibinfo{person}{Shuang Zhao}, {and} \bibinfo{person}{Henrik~Wann Jensen}.} \bibinfo{year}{2020}\natexlab{}.
\newblock \showarticletitle{A practical ply-based appearance model of woven fabrics}.
\newblock \bibinfo{journal}{\emph{ACM Trans. Graph.}} \bibinfo{volume}{39}, \bibinfo{number}{6}, Article \bibinfo{articleno}{251} (\bibinfo{date}{nov} \bibinfo{year}{2020}), \bibinfo{numpages}{13}~pages.
\newblock
\showISSN{0730-0301}
\urldef\tempurl%
\url{https://doi.org/10.1145/3414685.3417777}
\showDOI{\tempurl}


\bibitem[\protect\citeauthoryear{M\"{u}ller, Rousselle, Nov\'{a}k, and Keller}{M\"{u}ller et~al\mbox{.}}{2021}]%
        {mueller2021realtime}
\bibfield{author}{\bibinfo{person}{Thomas M\"{u}ller}, \bibinfo{person}{Fabrice Rousselle}, \bibinfo{person}{Jan Nov\'{a}k}, {and} \bibinfo{person}{Alexander Keller}.} \bibinfo{year}{2021}\natexlab{}.
\newblock \showarticletitle{Real-time Neural Radiance Caching for Path Tracing}.
\newblock \bibinfo{journal}{\emph{ACM Trans. Graph.}} \bibinfo{volume}{40}, \bibinfo{number}{4}, Article \bibinfo{articleno}{36} (\bibinfo{date}{Aug.} \bibinfo{year}{2021}), \bibinfo{numpages}{36:1--36:16}~pages.
\newblock
\urldef\tempurl%
\url{https://doi.org/10.1145/3450626.3459812}
\showDOI{\tempurl}


\bibitem[\protect\citeauthoryear{Olano and Baker}{Olano and Baker}{2010}]%
        {OlanoAndBaker:2010:Lean}
\bibfield{author}{\bibinfo{person}{Marc Olano} {and} \bibinfo{person}{Dan Baker}.} \bibinfo{year}{2010}\natexlab{}.
\newblock \showarticletitle{LEAN Mapping}. In \bibinfo{booktitle}{\emph{Proceedings of the 2010 ACM SIGGRAPH Symposium on Interactive 3D Graphics and Games}} \emph{(\bibinfo{series}{I3D '10})}. \bibinfo{publisher}{Association for Computing Machinery}, \bibinfo{address}{New York, NY, USA}, \bibinfo{pages}{181–188}.
\newblock
\showISBNx{9781605589398}
\urldef\tempurl%
\url{https://doi.org/10.1145/1730804.1730834}
\showDOI{\tempurl}


\bibitem[\protect\citeauthoryear{Sadeghi, Bisker, De~Deken, and Jensen}{Sadeghi et~al\mbox{.}}{2013}]%
        {Sadeghi:2013:Cloth}
\bibfield{author}{\bibinfo{person}{Iman Sadeghi}, \bibinfo{person}{Oleg Bisker}, \bibinfo{person}{Joachim De~Deken}, {and} \bibinfo{person}{Henrik~Wann Jensen}.} \bibinfo{year}{2013}\natexlab{}.
\newblock \showarticletitle{A practical microcylinder appearance model for cloth rendering}.
\newblock \bibinfo{journal}{\emph{ACM Trans. Graph.}} \bibinfo{volume}{32}, \bibinfo{number}{2}, Article \bibinfo{articleno}{14} (\bibinfo{date}{apr} \bibinfo{year}{2013}), \bibinfo{numpages}{12}~pages.
\newblock
\showISSN{0730-0301}
\urldef\tempurl%
\url{https://doi.org/10.1145/2451236.2451240}
\showDOI{\tempurl}


\bibitem[\protect\citeauthoryear{Wang, Jin, Ha\v{s}an, and Yan}{Wang et~al\mbox{.}}{2022}]%
        {Wang:2021:Sponge}
\bibfield{author}{\bibinfo{person}{Beibei Wang}, \bibinfo{person}{Wenhua Jin}, \bibinfo{person}{Milo\v{s} Ha\v{s}an}, {and} \bibinfo{person}{Ling-Qi Yan}.} \bibinfo{year}{2022}\natexlab{}.
\newblock \showarticletitle{SpongeCake: A Layered Microflake Surface Appearance Model}.
\newblock \bibinfo{journal}{\emph{ACM Trans. Graph.}} \bibinfo{volume}{42}, \bibinfo{number}{1}, Article \bibinfo{articleno}{8} (\bibinfo{date}{sep} \bibinfo{year}{2022}), \bibinfo{numpages}{16}~pages.
\newblock
\showISSN{0730-0301}
\urldef\tempurl%
\url{https://doi.org/10.1145/3546940}
\showDOI{\tempurl}


\bibitem[\protect\citeauthoryear{Williams}{Williams}{1983}]%
        {Williams1983PyramidalP}
\bibfield{author}{\bibinfo{person}{Lance Williams}.} \bibinfo{year}{1983}\natexlab{}.
\newblock \showarticletitle{Pyramidal parametrics}. In \bibinfo{booktitle}{\emph{Proceedings of the 10th Annual Conference on Computer Graphics and Interactive Techniques}} \emph{(\bibinfo{series}{SIGGRAPH '83})}. \bibinfo{publisher}{Association for Computing Machinery}, \bibinfo{address}{New York, NY, USA}, \bibinfo{pages}{1–11}.
\newblock
\showISBNx{0897911091}
\urldef\tempurl%
\url{https://doi.org/10.1145/800059.801126}
\showDOI{\tempurl}


\bibitem[\protect\citeauthoryear{Wu, Zhao, Yan, and Ramamoorthi}{Wu et~al\mbox{.}}{2019}]%
        {wu2019accurate}
\bibfield{author}{\bibinfo{person}{Lifan Wu}, \bibinfo{person}{Shuang Zhao}, \bibinfo{person}{Ling-Qi Yan}, {and} \bibinfo{person}{Ravi Ramamoorthi}.} \bibinfo{year}{2019}\natexlab{}.
\newblock \showarticletitle{Accurate appearance preserving prefiltering for rendering displacement-mapped surfaces}.
\newblock \bibinfo{journal}{\emph{ACM Trans. Graph.}} \bibinfo{volume}{38}, \bibinfo{number}{4}, Article \bibinfo{articleno}{137} (\bibinfo{date}{jul} \bibinfo{year}{2019}), \bibinfo{numpages}{14}~pages.
\newblock
\showISSN{0730-0301}
\urldef\tempurl%
\url{https://doi.org/10.1145/3306346.3322936}
\showDOI{\tempurl}


\bibitem[\protect\citeauthoryear{Xu, Wang, Zhao, and Bao}{Xu et~al\mbox{.}}{2017}]%
        {Xu:2017:LinearMip}
\bibfield{author}{\bibinfo{person}{Chao Xu}, \bibinfo{person}{Rui Wang}, \bibinfo{person}{Shuang Zhao}, {and} \bibinfo{person}{Hujun Bao}.} \bibinfo{year}{2017}\natexlab{}.
\newblock \showarticletitle{Real-Time Linear BRDF MIP-Mapping}.
\newblock \bibinfo{journal}{\emph{Computer Graphics Forum}} \bibinfo{volume}{36}, \bibinfo{number}{4} (\bibinfo{year}{2017}), \bibinfo{pages}{27--34}.
\newblock
\urldef\tempurl%
\url{https://doi.org/10.1111/cgf.13221}
\showDOI{\tempurl}


\bibitem[\protect\citeauthoryear{Zeltner, Rousselle, Weidlich, Clarberg, Nov\'{a}k, Bitterli, Evans, Davidovi\v{c}, Kallweit, and Lefohn}{Zeltner et~al\mbox{.}}{2024}]%
        {Zeltner2023RealTimeNA}
\bibfield{author}{\bibinfo{person}{Tizian Zeltner}, \bibinfo{person}{Fabrice Rousselle}, \bibinfo{person}{Andrea Weidlich}, \bibinfo{person}{Petrik Clarberg}, \bibinfo{person}{Jan Nov\'{a}k}, \bibinfo{person}{Benedikt Bitterli}, \bibinfo{person}{Alex Evans}, \bibinfo{person}{Tom\'{a}\v{s} Davidovi\v{c}}, \bibinfo{person}{Simon Kallweit}, {and} \bibinfo{person}{Aaron Lefohn}.} \bibinfo{year}{2024}\natexlab{}.
\newblock \showarticletitle{Real-Time Neural Appearance Models}.
\newblock \bibinfo{journal}{\emph{ACM Trans. Graph.}} (\bibinfo{date}{apr} \bibinfo{year}{2024}).
\newblock
\showISSN{0730-0301}
\urldef\tempurl%
\url{https://doi.org/10.1145/3659577}
\showDOI{\tempurl}
\newblock
\shownote{Just Accepted.}


\bibitem[\protect\citeauthoryear{Zhao, Jakob, Marschner, and Bala}{Zhao et~al\mbox{.}}{2011}]%
        {Zhao:2011:fabric}
\bibfield{author}{\bibinfo{person}{Shuang Zhao}, \bibinfo{person}{Wenzel Jakob}, \bibinfo{person}{Steve Marschner}, {and} \bibinfo{person}{Kavita Bala}.} \bibinfo{year}{2011}\natexlab{}.
\newblock \showarticletitle{Building volumetric appearance models of fabric using micro CT imaging}.
\newblock \bibinfo{journal}{\emph{ACM Trans. Graph.}} \bibinfo{volume}{30}, \bibinfo{number}{4}, Article \bibinfo{articleno}{44} (\bibinfo{date}{jul} \bibinfo{year}{2011}), \bibinfo{numpages}{10}~pages.
\newblock
\showISSN{0730-0301}
\urldef\tempurl%
\url{https://doi.org/10.1145/2010324.1964939}
\showDOI{\tempurl}


\bibitem[\protect\citeauthoryear{Zhao, Wu, Durand, and Ramamoorthi}{Zhao et~al\mbox{.}}{2016}]%
        {Zhao:2016:Anisotropic}
\bibfield{author}{\bibinfo{person}{Shuang Zhao}, \bibinfo{person}{Lifan Wu}, \bibinfo{person}{Fr\'{e}do Durand}, {and} \bibinfo{person}{Ravi Ramamoorthi}.} \bibinfo{year}{2016}\natexlab{}.
\newblock \showarticletitle{Downsampling scattering parameters for rendering anisotropic media}.
\newblock \bibinfo{journal}{\emph{ACM Trans. Graph.}} \bibinfo{volume}{35}, \bibinfo{number}{6}, Article \bibinfo{articleno}{166} (\bibinfo{date}{dec} \bibinfo{year}{2016}), \bibinfo{numpages}{11}~pages.
\newblock
\showISSN{0730-0301}
\urldef\tempurl%
\url{https://doi.org/10.1145/2980179.2980228}
\showDOI{\tempurl}


\bibitem[\protect\citeauthoryear{Zhu, Jarabo, Aliaga, Yan, and Chiang}{Zhu et~al\mbox{.}}{2023a}]%
        {Zhu:2023:cloth}
\bibfield{author}{\bibinfo{person}{Junqiu Zhu}, \bibinfo{person}{Adrian Jarabo}, \bibinfo{person}{Carlos Aliaga}, \bibinfo{person}{Ling-Qi Yan}, {and} \bibinfo{person}{Matt Jen-Yuan Chiang}.} \bibinfo{year}{2023}\natexlab{a}.
\newblock \showarticletitle{A Realistic Surface-Based Cloth Rendering Model}. In \bibinfo{booktitle}{\emph{ACM SIGGRAPH 2023 Conference Proceedings}} \emph{(\bibinfo{series}{SIGGRAPH '23})}. \bibinfo{publisher}{Association for Computing Machinery}, \bibinfo{address}{New York, NY, USA}, Article \bibinfo{articleno}{5}, \bibinfo{numpages}{9}~pages.
\newblock
\showISBNx{9798400701597}
\urldef\tempurl%
\url{https://doi.org/10.1145/3588432.3591554}
\showDOI{\tempurl}


\bibitem[\protect\citeauthoryear{Zhu, Montazeri, Aubry, Yan, and Weidlich}{Zhu et~al\mbox{.}}{2023b}]%
        {Zhu:2023:yarn}
\bibfield{author}{\bibinfo{person}{Junqiu Zhu}, \bibinfo{person}{Zahra Montazeri}, \bibinfo{person}{Jean-Marie Aubry}, \bibinfo{person}{Ling-Qi Yan}, {and} \bibinfo{person}{Andrea Weidlich}.} \bibinfo{year}{2023}\natexlab{b}.
\newblock \showarticletitle{A Practical and Hierarchical Yarn-based Shading Model for Cloth}.
\newblock \bibinfo{journal}{\emph{Computer Graphics Forum}} \bibinfo{volume}{42}, \bibinfo{number}{4} (\bibinfo{year}{2023}), \bibinfo{pages}{2--11}.
\newblock
\urldef\tempurl%
\url{https://doi.org/10.1111/cgf.14894}
\showDOI{\tempurl}


\end{thebibliography}

\mycfigure{compare_neumip}{Result_Compare_1_v8_compressed.pdf}{Comparison among our method, NeuMIP~\cite{Kuznetsov:2021:NeuMip} and modified Jin et al.~\shortcite{Jin:2022:inverse} on seen materials. While NeuMIP produces the lowest MSE occasionally, it has to be trained per material and does not support editing. The results by modified Jin et al.~\shortcite{Jin:2022:inverse} show noticeable aliasing, even with more samples. In contrast, our results are much smoother and free from aliasing.}

\mycfigure{compare_BRDF}{Result_Compare_2_v5.pdf}{Comparison between our method and modified Jin et al.~\shortcite{Jin:2022:inverse} on an unseen material (a twill $4 \times 4$ pattern). Our method produces results with lower MSE and less aliasing than both rendering results (1 SPP and 3 SPP) by modified Jin et al.~\shortcite{Jin:2022:inverse}.}

\mycfigure{compare_BTDF}{Result_Compare_3_v4.pdf}{Comparison between our method and modified Jin et al.~\shortcite{Jin:2022:inverse} on an unseen BTDF material (a plain pattern). Both rendering results (with 1 SPP and 3 SPP) by modified Jin et al.~\shortcite{Jin:2022:inverse} have apparent aliasing, while our results are closer to the ground truth.}

\mycfigure{ablation}{Result_Ablation_v8_compressed.pdf}{The influence of two designs (the one-blob encoding and the spatial fusion) in our network. With both components, our network has a more powerful representation ability, leading to a higher quality of rendering results.}

\mycfigure{ITP}{Result_ITP_v6_compressed.pdf}{
Interpolation and extrapolation of material latent vectors with roughness (left) and height field scaling factor (right) as the weights. The interpolated/extrapolated material latent vectors can produce rendered results similar to those obtained by the directly encoded material latent vectors.
}

\mycfigure{editing}{Result_Editing_v4.pdf}{Our method supports real-time editing of several parameters, including roughness, height field scaling factor, pattern (normal and orientation textures), and albedos, producing various rendering results.}

\end{document}